%% file: main.tex
\documentclass[11pt, a4paper, logo, copyright, nonumbering]{astribot}
\usepackage[numbers, sort&compress, square]{natbib}
\usepackage{dblfloatfix}
\usepackage{ulem}
\usepackage{caption}
\usepackage{xspace}
\usepackage{pifont}
\usepackage{multirow}
\usepackage{tcolorbox}
\usepackage{xltabular}
\usepackage{longtable}
\usepackage{hyperref}
\usepackage{amsfonts}
\usepackage{amsmath}
\usepackage{algorithm}
\usepackage[noend]{algpseudocode}
\usepackage{amssymb}
\usepackage{lineno}
\usepackage{adjustbox}
\usepackage{siunitx}
\usepackage{mathtools}
\usepackage[bottom]{footmisc}
\usepackage{fontawesome5}
\usepackage{subcaption}
\usepackage{setspace}
\usepackage{lipsum}
\usepackage{multicol}
\usepackage{pdfpages}
\usepackage{threeparttable}
\usepackage{makecell}
\usepackage{xcolor}
\usepackage{float}
\usepackage{tikz}
\usetikzlibrary{decorations.pathreplacing, patterns}

\tcbset{textmarker/.style={
        halign=center,parbox=false,boxrule=0mm,boxsep=0mm,arc=0mm,
        outer arc=0mm,left=1mm,right=1mm,top=7pt,bottom=7pt,
        toptitle=1mm,bottomtitle=1mm,oversize,}}

\newcommand{\model}[0]{\mbox{SmoothRL}\xspace}

\makeatletter
\def\@BTrule[#1]{%
  \ifx\longtable\undefined
    \let\@BTswitch\@BTnormal
  \else\ifx\hline\LT@hline
    \nobreak
    \let\@BTswitch\@BLTrule
  \else
     \let\@BTswitch\@BTnormal
  \fi\fi
  \global\@thisrulewidth=#1\relax
  \ifnum\@thisruleclass=\tw@\vskip\@aboverulesep\else
  \ifnum\@lastruleclass=\z@\vskip\@aboverulesep\else
  \ifnum\@lastruleclass=\@ne\vskip\doublerulesep\fi\fi\fi
  \@BTswitch}
\makeatother

\addto\extrasenglish{
}

 {\begin{list}{}%
         {\setlength{\leftmargin}{#1}}%
         \item[]%
 }
 {\end{list}}

\renewcommand{\today}{}

\title{\model: Online Reinforcement Learning During \\Asynchronous Execution}

\author[*]{
Astribot Team
\\
\small
\texttt{research@astribot.com}
\\
\vspace{2em}
\small
Project Page: \href{https://www.astribot.com/research/SmoothRL}{https://www.astribot.com/research/SmoothRL}
\\
\vspace{1em}
\small
Author List in \hyperref[sec:contribution]{Contributions}
}

\begin{abstract}
Deploying robot policies in the physical world requires simultaneously satisfying two fundamental desiderata: reliability, as reflected by high task success rates, and smooth real-time execution. However, deploying state-of-the-art generalist models, including Vision-Language-Action (VLA) models and World-Action Models (WAMs), presents significant challenges on both fronts. Achieving the precision and robustness required for real-world deployment necessitates sample-efficient online reinforcement learning (RL) to adapt pretrained models for reliable deployment. Meanwhile, the increasing scale of robot foundation models has led to correspondingly higher inference latency. To satisfy real-time constraints under high latency, modern systems adopt asynchronous inference together with action chunking, overlapping policy computation with chunk execution to hide latency and enable smooth control. Despite their complementary roles, integrating asynchronous execution with gradient-based online RL remains underexplored.
We present \model, an online RL framework that fine-tunes a pretrained policy within an asynchronous inference loop. \model follows a value-gradient paradigm, directly updating the policy parameters using gradients of the action-value function with respect to the policy actions. To enable correct optimization under asynchronous execution, \model explicitly models the asynchronous inference process during training. Specifically, each generated action chunk is partitioned by frame index into three regions: a \textit{committed region}, which consists of actions committed by the previous inference cycle; an \textit{execution region}, which contains newly generated actions that are executed by the robot; and a \textit{discarded region}, which contains actions that will be superseded by the next inference cycle. Gradients are propagated only through the execution region, ensuring that policy optimization is aligned with the trajectory distribution induced by asynchronous execution.
We evaluate \model on real-world robotic tasks requiring high precision, as well as highly dynamic tasks that necessitate asynchronous execution.
\end{abstract}

\begin{document}
\thispagestyle{firststyle}
\maketitle
\clearpage

\pagestyle{plain}
\input{sessions/introduction}
\input{sessions/related_work}

\input{sessions/method}

\input{sessions/experiments}
\input{sessions/conclusion}

\section{Contributions}
\label{sec:contribution}
\begin{itemize}
    \item  \textbf{Contributors:} Guang Gao$^*$, Yuxuan Nong$^*$, Baifu Huang
    \item  \textbf{Project Lead:} Jianan Wang
\end{itemize}

\noindent{\small $^*$ Equal contribution.}
\clearpage

\input{sessions/bibliography}
\end{document}

%% file: sessions/introduction.tex
\input{sessions/fig_teaser}
\section{Introduction}
\label{sec:intro}

Recent advances in Vision-Language-Action (VLA) models and World-Action Models (WAMs) have demonstrated remarkable progress toward endowing robots with general manipulation capabilities~\cite{pi0,rdt,groot,fastwam,cosmos,lingbot,beingh07,lumo2,wog}. However, deploying such generalist policies in the physical world requires simultaneously addressing two fundamental challenges: \textbf{reliability} and \textbf{smooth real-time execution}.
Achieving the robustness required for diverse real-world tasks such as cutting open sealed packages demands sample-efficient online reinforcement learning (RL). Meanwhile, the increasing scale of robot foundation models has led to correspondingly higher inference latency. Under synchronous execution, this latency lies directly on the control path: the robot must pause at every chunk boundary while waiting for the next forward pass to complete. When success depends on maintaining continuous motion, such pauses can severely degrade both task success and execution efficiency. Modern robot systems therefore adopt action chunking together with asynchronous inference, overlapping policy inference with action execution.
Asynchronous execution eliminates these pauses but introduces a discontinuity of its own: each incoming chunk is generated from an earlier observation, so switching to it can produce a discontinuity at the junction. A family of chunk-stitching methods~\cite{rtc,ttrtc,legato,softrtc} therefore studies how to maintain continuity between newly generated actions and those already committed for execution.
Although reliability and smooth real-time execution are complementary requirements for real-world deployment, they have largely been studied in isolation. Existing online RL methods typically optimize policies assuming synchronous execution, while deployment almost invariably relies on asynchronous inference. This mismatch causes the policy to be optimized under dynamics that differ from those encountered at test time. We argue that online RL should instead be performed under the same asynchronous execution paradigm used during deployment: \textbf{Reinforce in Deployment.}

Pursuing reliability and smooth real-time execution simultaneously reduces the design space to three essential requirements. \textbf{Online}: the policy improves from the data it is currently generating, rather than through offline training rounds. \textbf{Asynchronous}: inference overlaps execution so that model inference latency does not limit the control frequency. \textbf{Value-gradient}: the gradient of the action-value function with respect to the policy actions is backpropagated into the policy parameters, allowing the learning signal to directly optimize the policy. Existing approaches each sacrifice one of these requirements, in the same order.
\textbf{Offline RL} separates learning from deployment by conducting RL in offline rounds, leaving deployment free to be asynchronous. Estimated advantages influence policy updates through reweighting or conditioning, rather than direct gradient optimization through the value function~\citep{chi0,recap}.
\textbf{Synchronous Online RL} performs online reinforcement learning by backpropagating gradients from the value function $Q$ into the policy parameters. While effective for adapting pretrained policies~\citep{rlt,expoft,pld,residualoprl,dawn}, these methods assume synchronous execution.
\textbf{Asynchronous Restricted Online RL} considers online adaptation under asynchronous execution. However, policy optimization is confined to a latent steering space rather than the policy parameters themselves, making the achievable improvement ultimately bounded by the decoding capability of the pretrained base policy~\citep{grrl}.
\textbf{Asynchronous Online RL}, which satisfies all three requirements, is the focus of this work. Under asynchronous execution, each generated action chunk is only partially executed before a newly inferred chunk supersedes its remaining actions. However, policy updates should be driven exclusively by the consequences of the actions the robot actually executes. Unlike the three settings above, asynchronous online RL cannot avoid gradient contamination by design: taking the value-gradient pathway while deploying asynchronously inevitably exposes the policy parameter update to un-executed, superseded actions. The objective itself must therefore eliminate this contamination, rather than relying on the execution schedule.

We present \model, a general framework for online RL fine-tuning of pretrained robot policies under asynchronous inference, enabling highly dexterous and dynamic tasks, such as throwing an object at a target, as illustrated in Figure~\ref{fig:teaser}. The framework follows the value-gradient paradigm, where the action-value function $Q$ updates the policy parameters directly through its gradient with respect to the policy actions.
This paradigm admits either fine-tuning the pretrained base policy directly or learning a lightweight policy on top of a frozen one; we adopt the latter as one instantiation, where a frozen base policy generates reference action chunks under a receding horizon and a trainable attached network in the raw action space carries all learnable parameters.
Under asynchronous execution, each generated action chunk is partitioned according to its execution status into three regions: a \emph{committed region}, containing actions already issued during the previous inference cycle; an \emph{execution region}, containing newly generated actions that are actually executed by the robot; and a \emph{discarded region}, containing newly generated actions that are superseded by the next inference cycle before execution. The boundaries between these regions are determined by two distinct forward passes and are unknown at chunk generation time, unless the inference loop is constrained to a fixed latency budget. Moreover, because inference and execution overlap, a chunk begins execution only after the preceding in-flight inference has completed, making each decision inherently concurrent~\cite{concurrent}: the action chunk currently in flight is part of the state against which the next chunk is generated. The two contributions below formalize the resulting implications for training:

\begin{itemize}
  \item \textbf{Gradient truncation to the execution region.}
  Holding the asynchronous loop to a latency budget of $n$ frames fixes both boundaries of the execution window, making the three action-chunk regions a function of frame index alone. The value gradient $\nabla_a Q$ is then computed only with respect to the execution region $a_{[n,2n)}$, ensuring that policy updates depend exclusively on actions whose consequences are observed. The critic, however, retains the committed region as part of its input for two reasons: the chunk-skip bootstrap is unbiased only when the critic is conditioned on the action sequence that generated the interval reward, while the action currently in flight provides the state augmentation required to model the concurrent decision process.
  \item \textbf{Asynchronous execution embedded in the training loop.}
    The asynchronous inference loop - where policy forward passes and action execution overlap in time - runs during training rollouts and not only at deployment. Replay trajectories therefore record the actions the robot actually executed under the same timing and execution schedule the objective is defined against, ensuring the policy is never optimized under dynamics it will not encounter at deployment.
\end{itemize}

A further benefit of operating in the raw action space is that human interventions lie in the same space as the policy outputs. Consequently, a demonstrated chunk can be used directly, without latent inversion or a separate offline data-collection stage: it serves as a regression target for the policy under a behavioral cloning loss over the execution region and as a transition for the critic under the standard TD backup.
We evaluate \model on a suite of real-world robotic manipulation tasks requiring high precision, as well as highly dynamic tasks that necessitate asynchronous execution.

%% file: sessions/fig_teaser.tex
\begin{figure}[t]
\centering
\includegraphics[width=\linewidth]{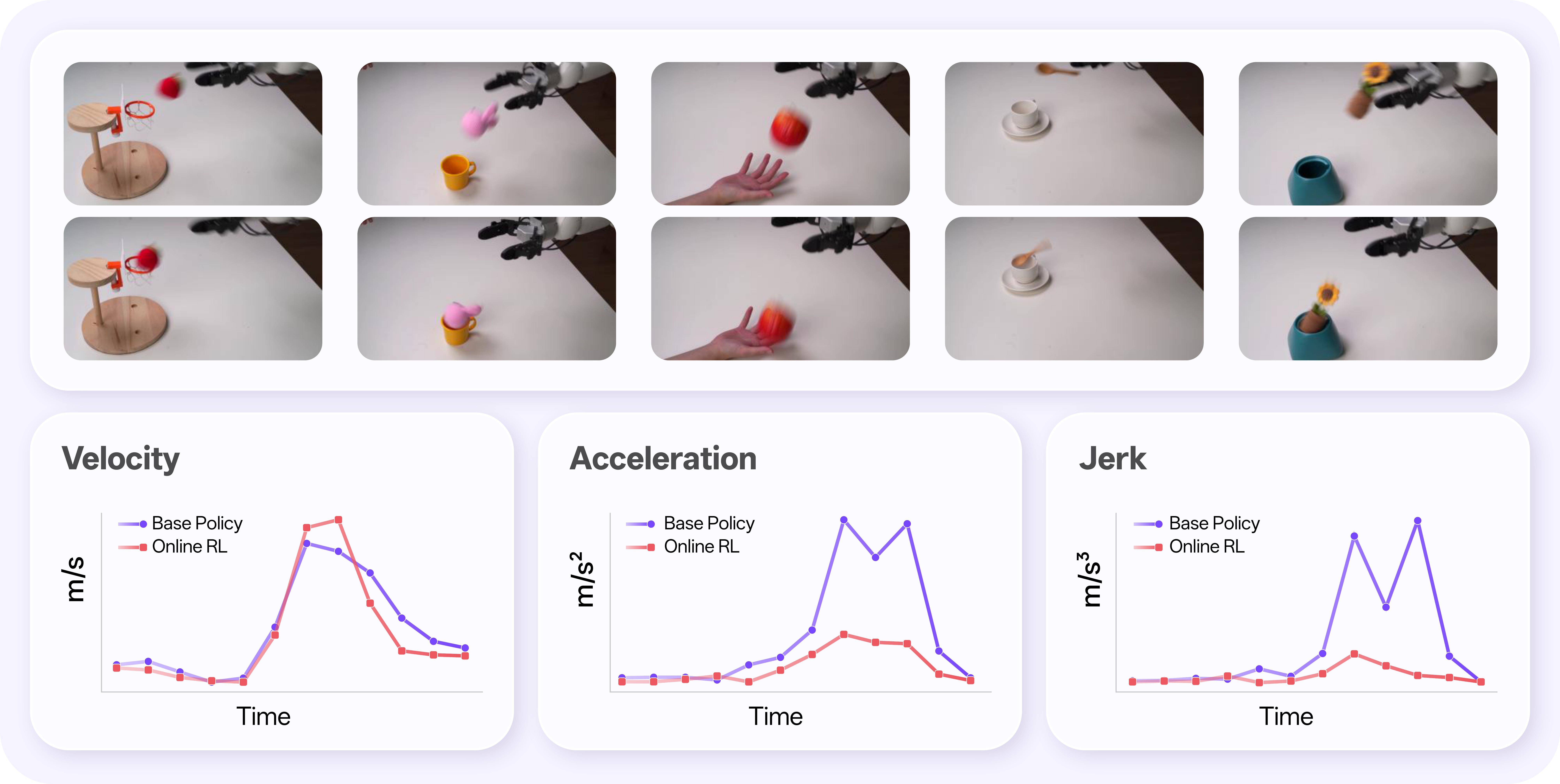}
\caption{
\textbf{Top:} \model enables online RL fine-tuning of pretrained robot policies under asynchronous inference, supporting highly dexterous and dynamic tasks such as throwing an object at a target. 
\textbf{Bottom:} \model produces smoother motion, reducing acceleration and jerk by 52\% and 47\%, respectively. Values are RMS measurements of the right end-effector's XYZ motion over each action chunk during a real autonomous throwing rollout.
}
\label{fig:teaser}
\end{figure}

%% file: sessions/related_work.tex
\section{Related Work}
\label{sec:rw}

We first review mechanisms for asynchronous policy deployment, and then organize existing RL approaches along the three dimensions introduced in \S\ref{sec:intro}: whether learning is online or offline, whether deployment is synchronous or asynchronous, and whether value gradients are propagated directly to the policy.

\subsection{Asynchronous Policy Deployment}
\label{sec:rw1}
Receding-horizon action chunking has become the standard deployment paradigm for contemporary large-scale generalist robot policies~\cite{dp,act,pi0,rdt,groot,fastwam}. Since generating each action chunk requires a complete policy forward pass, typically taking tens to hundreds of milliseconds, asynchronous inference is a practical necessity: synchronous per-segment forwards would pin control frequency to an unacceptable rate~\cite{asyncsurvey,vlash,duocore}. Consequently, each newly generated chunk is conditioned on a stale observation and is unaware of the actions executed since that observation was captured. As a result, successive chunks generally fail to connect smoothly along the executed trajectory, introducing velocity or acceleration discontinuities at chunk boundaries.
One line of work addresses this problem through \textbf{consistency-aware chunk-stitching }~\cite{rtc,ttrtc,legato,softrtc}. These methods constrain the already-consumed prefix of each newly generated chunk to match the actions that have already been executed, while leaving the remaining horizon unconstrained. A related approach instead corrects an in-flight chunk using newly acquired observations~\cite{a2c2}. Over time, these methods have evolved from inference-time guidance external to the policy toward training-time modeling within the policy, and from hard prefix constraints toward soft interpolation.
An orthogonal line of work employs \textbf{trajectory blending}~\cite{act,chi0}, which smooths chunk boundaries by post-processing the overlapping predictions of successive chunks using a decaying interpolation weight. Rather than constraining the policy outputs themselves, trajectory blending operates only after multiple chunks have been generated. Although both approaches eliminate discontinuities, they differ fundamentally in when the executed action is determined: chunk stitching fixes the action before it is emitted by the policy, whereas trajectory blending determines it only after the chunks overlapping that instant become available.
This distinction is inconsequential for imitation learning and offline RL, where both families have primarily been studied. However, it becomes decisive for the online RL objective, which requires the action at which the value function $Q$ is evaluated to be exactly the action executed by the robot, while simultaneously preserving a differentiable path from that action to the policy parameters. This requirement raises a new question for asynchronous online RL, which we address in this work. 

\subsection{Offline RL}
\label{sec:rw2b}
Offline RL consumes the learning signal entirely during training, leaving the policy fixed at deployment. Methods such as $\chi_0$~\cite{chi0} and RECAP~\cite{recap} support asynchronous deployment: the advantage is estimated by a separate value model and influences the policy only through weighting (AWR~\cite{awr}) or conditioning (CFG-style guidance~\cite{cfgrl}), rather than by backpropagating value gradients into the policy parameters. Conversely, CO-RFT~\cite{corft} performs all chunk-level $Q$-learning offline (Chunked Cal-QL) but adopts synchronous deployment. The coexistence of both deployment modes within this class shows that, for offline RL, asynchronous execution is merely an inference-scheduling choice and does not alter the learning signal. In contrast, our method closes the actor-critic loop around real-robot execution, where the mismatch between generated and executed actions directly affects the learning signal.

\subsection{Synchronous Online RL}
\label{sec:rw2a}
This body of work performs online actor-critic updates in the same loop as robot rollout. All methods in this category follow the value-gradient paradigm, in which the action-value function backpropagates gradients directly into the policy parameters. They differ primarily in where reinforcement learning is applied: either on a lightweight module attached to a frozen pretrained policy, or on the entire policy through end-to-end fine-tuning.
One line of work freezes the pretrained VLA and learns a chunk-level policy on top. RLT~\cite{rlt}, whose architecture our implementation directly adopts, appends a learnable RL token to a frozen VLA and conditions a chunk-level actor on both the resulting representation and the VLA reference action. The actor is optimized with a TD3-style~\cite{td3} off-policy objective, while a behavior-cloning regularizer constrains learning to local corrections. EXPO-FT~\cite{expoft} similarly attaches a residual editing head to a frozen VLA and optimizes it using a SAC-style~\cite{sac} actor objective together with Best-of-$N$ selection, placing it within the same value-gradient framework. Other methods retain the frozen-base interface but learn per-step residual policies rather than chunk-level ones~\cite{pld,residualoprl,dawn,expo,hilserl}, reducing the RL optimization unit from an action chunk to a single control step. Chunk-level off-policy methods outside the frozen-policy setting~\cite{toperl,taxonomy} nevertheless adopt the same chunk-level optimization unit as our approach. This shared optimization unit motivates our formalization of the \textbf{action chunk as the fundamental RL optimization unit} (\S\ref{sec:m2}). In a per-step formulation, actions carry no frame index within a chunk, making it impossible to express gradients that apply only to a selected sub-segment of that chunk.
A second line of work applies reinforcement learning directly to the entire policy. Representative examples include end-to-end token-level optimization for VLAs~\cite{vlarl,simplevlarl,irevla,acppo}, reinforcement learning of flow- and diffusion-based policies~\cite{pirl,pistepnft,dppo,reinflow,sernf}, and consistency-distillation approaches~\cite{rl100,conrft}. FQL~\cite{fql}, while operating in an offline-to-online rather than closed-loop real-robot setting, distills a pretrained flow policy into a one-step policy that directly receives value gradients, a design also explored as a variant of QAM~\cite{qam}. LWD~\cite{lwd} closes the loop on real-robot execution by performing fleet-scale online updates with a chunk-level critic and propagating $\nabla_a Q$ into a flow-policy action head through an adjoint-matching formulation.
Despite their differences, all methods in this category assume synchronous receding-horizon execution, so partial chunk execution never enters their formulation. By contrast, opening the deployment loop to asynchronous inference turns partial execution into a constraint on the value-gradient pathway. Our method addresses this constraint through chunk-role partitioning and gradient truncation, both defined over frame indices within an action chunk. Because the formulation operates directly on raw action chunks rather than a particular policy parameterization, it applies equally to residual-policy adaptation and end-to-end policy fine-tuning.

\subsection{Asynchronous Restricted Online RL}
\label{sec:rw2c}
Online actor-critic learning requires exact chunk transitions, yet asynchronous execution causes each generated action chunk to be only partially executed, contaminating those transitions. Existing methods avoid this issue by keeping the value gradient, $\nabla_a Q$, away from the raw-action policy parameters. Instead, reinforcement learning is performed in latent space, through value-based action selection, or by using the value function only to rank or verify sampled candidates~\cite{vgps,robomonkey}. Our method instead follows the value-gradient paradigm, embedding asynchronous execution directly into the RL objective as a known optimization constraint.
GR-RL~\cite{grrl} performs online learning under asynchronous execution, but places the RL optimization variable in the latent noise space of a frozen diffusion policy~\cite{dsrl}. It learns a latent-noise predictor, while a frozen decoder maps the optimized latent back to actions. Optimizing the latent noise therefore selects among actions already represented by the pretrained policy prior. Consequently, partial execution is bounded in its effect rather than eliminated: if value estimation includes un-executed portions of an action chunk, the resulting error may alter which latent candidate is selected, but the frozen decoder ensures that every issued action remains within the action distribution represented by the pretrained prior. This property bounds not only the error but also the attainable improvement, which remains limited by the expressive capacity of the frozen decoder. Moreover, human intervention cannot be incorporated as direct policy regression targets without unreliable latent inversion. In contrast, our method optimizes directly in raw action space, allowing human intervention data to serve simultaneously as supervised regression targets and value-labeled online transitions.

%% file: sessions/method.tex
\section{Method}
\label{sec:method}

Our work addresses one single challenge: 
\begin{center}
\emph{How can online RL fine-tuning be integrated into an asynchronous deployment loop ?}
\end{center}
\subsection{The Asynchronous Inference Loop}
\label{sec:m1}
We exclusively consider an asynchronous inference setting to satisfy the smooth real-time execution requirements of real-world deployment, which has become the standard paradigm for contemporary large-scale generalist robot policies. The base policy infers the next action chunk while the robot continues executing the current one. Meanwhile, the control process operates at the desired control frequency and always issues the latest available action without waiting for inference to complete; once a new chunk becomes available, it immediately takes over execution. This design effectively hides inference latency.

Asynchronous execution introduces a fundamental consequence: \textbf{a policy-generated action chunk is only partially executed by the environment.} We therefore partition each action chunk of horizon $H$ into three regions by frame index, according to how the asynchronous schedule uses the generated actions. The \textbf{committed region} $[0,d)$ contains the frames that elapse during the inference latency of this chunk. Its length $d$ is determined by the current inference time. By the time the chunk becomes available, these frames have already passed: the previous chunk has supplied the corresponding actions, and the predictions in this region can never affect the environment through the current chunk. The \textbf{execution region} $[d,d+c)$ corresponds to the interval during which this chunk is the latest available policy output before the next inference completes. Its length $c$ is determined by the arrival time of the subsequent chunk. Every action in this region is issued to the robot exactly as generated, making it the only part of the chunk that directly determines the resulting environment trajectory. The remaining \textbf{discarded region} contains frames after the next chunk becomes available. These actions are superseded by the newly arrived chunk and therefore never reach the environment. 

\input{sessions/fig_chunk_partitioning}

\paragraph{Stabilizing the loop with a latency budget.}
The boundaries between chunk regions, illustrated in Figure~\ref{fig:chunk_partitioning} (a), are stochastic in real deployments. Since they are determined by different inference cycles, they fluctuate independently and are unknown when an action chunk is generated.
Rather than modeling this stochasticity, we eliminate it through a fixed latency budget. We estimate an upper bound $n$ on the inference time and enforce a scheduled execution loop: each chunk is requested $n$ control steps before its first frame is due. If inference completes earlier, the system waits, ensuring that chunk handover always occurs at the predefined time. Consequently, both region lengths are fixed by the same constant, $d=c=n$. For any horizon $H \geq 2n$, the committed region is therefore $[0,n)$, the execution region is $[n,2n)$, and the discarded region is $[2n,H)$ (Figure~\ref{fig:chunk_partitioning} (b)). The slack between the actual inference latency and the budget $n$ is converted into waiting time rather than used to exploit fresher observations. In exchange, every chunk is executed over an identical window determined solely by its frame indices.

\subsection{Gradient-based RL in Continuous Action Space}
\label{sec:m2}

This subsection reviews the general formulation of continuous-action reinforcement learning used throughout the paper. Given a Markov decision process (MDP), a policy $\pi$ seeks to maximize the expected discounted return. Value-based methods define the action-value function $Q^\pi(s,a)$, which satisfies the Bellman equation:
\begin{equation}
Q^\pi(s, a) = \mathbb{E}_{r,\, s' \sim P}\big[\, r + \gamma\, Q^\pi(s', \pi(s')) \,\big],
\end{equation}
where $\gamma$ is the discount factor and $s'$ is the successor state reached after executing action $a$ in state $s$. Off-policy actor-critic methods alternate between two updates. The \textbf{critic} learns the action-value function $Q$ by fitting the temporal-difference (TD) target given by the right-hand side of the Bellman equation using transitions sampled from a \textbf{replay buffer}. The \textbf{actor} then updates the policy parameters to increase the predicted action value, i.e., to maximize $Q$.

In continuous action spaces, maximizing the action value can be realized through two distinct routes. One route improves actions through candidate selection, advantage weighting, or latent-space guidance, without backpropagating $\nabla_a Q$ to the policy parameters that generate the executed actions. Our framework instead follows the value-gradient route: the gradient of the action-value function with respect to the action, $\nabla_a Q$, is backpropagated through the deterministic policy $a=\pi_\theta(s)$ to the policy parameters $\theta$,
\begin{equation}
\nabla_\theta J = \mathbb{E}\big[\, \nabla_a Q(s, a)\big|_{a = \pi_\theta(s)} \cdot \nabla_\theta \pi_\theta(s) \,\big],
\end{equation}
so that $Q$ directly updates the policy through action-space gradients.

\paragraph{Action Chunk as the Atomic Unit of RL.}
We formulate reinforcement learning over action chunks, treating each chunk as an atomic action in the underlying MDP. For a chunk spanning $H$ frames, the transition is represented as $(s_t, a_{t:t+H}, r_t, s_{t+H})$, where $a_{t:t+H}$ is the action chunk generated by the policy at state $s_t$, $r_t$ is the discounted return accumulated during execution of the chunk, and $s_{t+H}$ is the observation after the chunk has been fully executed. When indexing frames relative to the start of a chunk rather than to the episode - as in the chunk regions defined in \S\ref{sec:m1} - we denote the chunk by $a_{[0,H)}$ and a sub-segment by $a_{[i,j)}$. The corresponding action-value function is written $Q(s_t, a_{t:t+H})$, and its temporal-difference update follows the chunk-skip Bellman backup:
\begin{equation}
Q(s_t, a_{t:t+H}) \;\leftarrow\; r_t + \gamma^{H}\, Q\big(s_{t+H},\ \pi(s_{t+H})\big).
\label{eq:backup}
\end{equation}
This backup remains unbiased because the critic is conditioned on the action sequence that generated the reward over the corresponding interval. The same formulation has since been adopted by recent chunk-level RL methods (\S\ref{sec:rw2a}), where the action span on which the critic is conditioned need not coincide with the span for which the policy is optimized~\cite{dqc}. Our framework likewise adopts this formulation, with the modification required to account for asynchronous execution.

\subsection{Chunk-Level RL under Asynchronous Deployment}
\label{sec:m3}

Under asynchronous deployment, directly concatenating the two components above - allowing $\nabla_a Q$ to backpropagate through the entire chunk - would cause policy updates to depend on generated actions that never reached the environment. Instead, we explicitly treat ``which segment is actually executed'' as a known constraint imposed by the deployment process.

The regions introduced in \S\ref{sec:m1} were originally defined solely by the execution schedule; here, they are assigned distinct training roles. From this point onward, the span over which the objective is defined is restricted to the first $2n$ frames of the chunk, with the committed and execution regions of \S\ref{sec:m1} each occupying one budget interval; the horizon $H$ the base policy emits may be longer. We denote by $\tilde a_{[0,n)}$ the actions actually executed by the robot in the committed region of the current chunk. These actions were issued by the preceding chunk and are treated as fixed, regardless of how they were generated. The objective defined in this subsection is therefore:

\begin{equation}
\max_{\theta}\ \ \mathbb{E}\Big[\, Q\big(s,\ \tilde a_{[0,n)},\ \pi_\theta(s)_{[n,2n)}\big) \Big]
\quad\text{s.t.}\quad
(\tilde a_{[0,n)},\ \pi_\theta(s)_{[n,2n)}) \in \mathcal{E}
\label{eq:objective}
\end{equation}

where $\mathcal{E}$ denotes the set of chunks over $[0,2n)$ that are physically executable by the robot. Importantly, admissibility is a property of the full chunk span, not merely the execution region: a chunk is included only when its execution is valid within both regions and remains consistent across the boundary at frame $n$. The discarded region $[2n,H)$, which falls outside the effective execution horizon, is therefore excluded entirely from the objective above. The chunk-skip backup of Eq.~\eqref{eq:backup} is restricted to the same span: the interval it spans is $[0,2n)$ rather than the full horizon $H$, so the reward $r_t$ accumulates over those $2n$ frames, the bootstrap state is $s_{t+2n}$, and the discount exponent is $2n$. This modification preserves the required unbiased backup: the critic is conditioned on the action sequence that actually generated the reward over the interval, namely $\tilde a_{[0,n)}$ followed by $a_{[n,2n)}$.

\paragraph{Value gradient truncation.}
From the policy's perspective, the committed region induces no valid environmental feedback for policy updates: as described in \S\ref{sec:m1}, these actions are overwritten by those issued from the preceding chunk and therefore do not form a valid gradient path for the current policy. Accordingly, the value gradient is truncated to the execution region: $\nabla_a Q$ is computed only with respect to $a_{[n,2n)}$. This truncation establishes a crucial identity: the segment receiving the value gradient is simultaneously the policy output and the action actually executed by the environment, making the resulting gradient well-defined with respect to the policy parameters. Importantly, the truncation applies only to the gradient path, while the critic continues to operate on the full action chunk.

\paragraph{Critic over the full action span.} Asynchronous execution turns this setting into a concurrent decision problem. The execution region does not begin acting from $s_t$, but from $s_{t+n}$, with the intervening frames being governed by the chunk already in flight. A critic conditioned only on $(s_t, a_{[n,2n)})$ would therefore marginalize over the unknown distribution of in-flight chunks present in the replay buffer, which, in the off-policy setting, reflects behaviors induced by older policies. A concurrent decision process requires augmenting the state with two quantities~\cite{concurrent}: the action currently in flight and the remaining time until its completion. In our setting, the latency budget fixes the latter to the constant $n$, while chunk-stitching consistency makes the former directly available at generation time: for each newly generated chunk, its committed region exactly corresponds to the actions already issued by the preceding chunk. Recording these issued actions therefore provides the required state augmentation both freely and exactly. The same latency budget that determines this augmentation also fixes the timescale of the bootstrap chain, which we analyze further in \S\ref{sec:conclusion}.

\paragraph{Smoothness.}

The executable-set constraint restricts not the frames over which the policy can act, but the set of chunks over which the optimization is performed. The actor also conditions on the committed region, but for a different reason than the critic: since the committed and execution regions are connected at frame index $n$, a candidate chunk must be selected with awareness of the trajectory it must continue from. Smoothness inherited from the pretrained policy is provided by the demonstrations, rather than explicitly enforced. However, once the value objective increases $Q$ over the execution region, this smoothness is no longer guaranteed. Continuity must therefore be reintroduced as an explicit constraint in the objective. The optimization consequently searches only over $\mathcal{E}$, transferring the requirement of smoothness back to the policy.
The formulation above is defined with respect to the timing model in \S\ref{sec:m1}; therefore, transitions must be collected under the same execution semantics. The asynchronous loop runs during training rollouts rather than only at deployment, ensuring that the replay buffer records the actions actually executed by the robot. In contrast, transitions collected under synchronous execution would contain no explicit region boundaries for the objective to align with.

Exploration is fundamental to reinforcement learning. However, exploration on a real robot is costly: failed trials consume wall-clock time and, for contact-rich tasks, may incur hardware wear or damage. Consequently, human intervention during impending policy failures has become a standard component of real-robot RL rather than merely a convenience~\cite{hilserl,rlt,expoft,recap}. The objective above naturally accommodates such interventions without modification. The critic takes a state and an action chunk as input, and the target in Eq.~\eqref{eq:backup} is independent of the controller that generated the chunk; thus, intervened episodes can be stored in the replay buffer as ordinary transitions. The state augmentation introduced above remains valid under intervention: $\tilde a_{[0,n)}$ is defined by the actions actually issued rather than the actions proposed by the policy, and therefore directly records human commands when intervention occurs. Interpreting human actions as value-estimated actions requires no additional transformation in our formulation, unlike settings where the RL variable is represented in a latent action space (\S\ref{sec:rw2c}). \S\ref{sec:m4} describes the corresponding intervention mechanism.

Under asynchronous deployment, two actions must be distinguished: the action emitted by the policy and the action executed by the robot. Optimizing $Q$ with respect to the former improves behavior only when the two coincide. Furthermore, the action at which $Q$ is evaluated must admit a valid gradient path back to $\theta$. We refer to these two requirements as \emph{agreement} and \emph{differentiability}, respectively. Issuing the policy output directly satisfies both by construction, which is the setting assumed in Eq.~\eqref{eq:objective}; this is precisely where the two boundary-smoothing families discussed in \S\ref{sec:rw1} diverge. A consistency constraint modifies the chunk before execution, ensuring that the policy output and the executed action remain identical. In contrast, blending is performed only after both the new and previous chunks become available. To preserve the agreement requirement, the critic must evaluate the blended action actually executed by the robot. The truncation principle introduced above - that gradients propagate only through actions that enter the environment - extends naturally to this setting. The value gradient should be computed with respect to every chunk contributing to the executed window, restricted to the segment that the chunk contributes. Because these segments are disjoint across transitions, each chunk is differentiated exactly once. This extension requires an objective horizon exceeding $2n$, an additional actor forward pass for each contributing chunk to recover its output under the current policy parameters, and a reformulation of the continuity constraints around the mixing operator (\S\ref{sec:conclusion}). For simplicity, we adopt direct policy output execution and enforce smoothness through constrained optimization.

\input{sessions/fig_architecture}

\subsection{Instantiation}
\label{sec:m4}

We provide a concrete instantiation of the proposed asynchronous online RL framework, as illustrated in Figure~\ref{fig:architecture}. Our implementation follows the RLT~\cite{rlt} skeleton: a frozen base policy generates reference action chunks, while a lightweight TD3-style actor-critic accesses the base policy's internal representations through an RL token and predicts a residual action correction. The following paragraphs clarify what this skeleton provides and where our implementation departs from it.
The skeleton requires no structural modification to satisfy the requirements of \S\ref{sec:m3} : it provides a chunk-level actor, making frame-indexed gradient truncation expressible; operates in the raw action space, eliminating the need to invert human interventions; and exposes a critic input to which the committed region can be appended as a stop-gradient term. Keeping the base policy frozen and confining all trainable parameters to a lightweight attached network also makes each gradient step inexpensive relative to a base-policy forward pass. This is important because the update rate is bounded by rollout throughput, while the small number of trainable parameters makes online optimization practical at the data volume generated by a real robot.
The choices that follow are therefore implementation details rather than requirements of the framework: \S\ref{sec:m1}--\S\ref{sec:m3} are formulated in terms of generic $\pi$ and $Q$, and none of their arguments depends on this particular instantiation.

\input{sessions/fig_intervention}

\paragraph{Human intervention.}
To evolve the policy during deployment, we support two human intervention modes, as illustrated in Figure~\ref{fig:intervention}: \textbf{absolute intervention} and \textbf{residual intervention}. Intervention is activated and deactivated by a switch on the teleoperation device. The training loop continues while the human is in control: both the base policy and the attached network continue to perform inference, while only the issued action is modified. Under absolute intervention, the teleoperated chunk is issued directly and the actor output is discarded. Under residual intervention, the human input is added to the actor output, allowing the operator to modify the policy-generated chunk rather than replace it. Each mode is supported by a suitable input device: absolute intervention uses VR teleoperation, where the operator's hand pose directly specifies the target action, while residual intervention uses a hand controller, where stick displacements specify action deltas.

The two intervention modes are suited to different task regimes. Residual intervention preserves the temporal structure of the underlying action chunk while modifying its trajectory, allowing the operator to correct the policy without disrupting its velocity profile. We therefore use it for dynamic tasks. Absolute intervention, in contrast, gives the operator full control over the action chunk and is therefore suited to tasks requiring greater flexibility and high precision, where the demonstrated motion may differ substantially from the trajectory proposed by the policy.
Despite their different interaction mechanisms, the two modes are treated identically downstream. The action actually issued by the operator is recorded in $\tilde a_{[0,n)}$ and used as the BC target. Chunks are flagged as intervened at the chunk level, and this flag determines the BC target for the execution region: $a^{\text{target}}_{[n,2n)}$ is the action chunk actually issued for an intervened chunk and the base-policy reference otherwise, while the committed region always uses the reference action as its target. The input to $Q$ is independent of the intervention flag: because the committed region records the action actually issued, it contains the human actions whenever the preceding chunk was intervened. Thus, the state augmentation introduced in \S\ref{sec:m3} continues to faithfully represent the action actually in flight.

\paragraph{The base-policy interface.}
The skeleton's interface is preserved: the trainable networks access the base policy through a compressed token of its internal representation, while the actor is conditioned on the reference action chunk generated by the base policy. Let $\pi_{\text{base}}$ denote the frozen base policy and $E$ the encoder that reads its internal representation during chunk inference and compresses it into the RL token $z_t = E(\pi_{\text{base}})$. 
The base policy runs under a TT-RTC scheduler~\cite{ttrtc}to maintain the timing guarantees of \S\ref{sec:m1}.
Each base-policy forward pass therefore produces both the reference action chunk $\bar a_{t:t+2n} = \pi_{\text{base}}(s_t)$ and the RL token $z_t$ on which the trainable networks are conditioned. The committed region provided to $Q$ remains $\tilde a_{[0,n)}$ as defined in \S\ref{sec:m3}, which coincides with the reference chunk only in the absence of intervention.
For the concrete instantiation, the state argument of the generic $\pi$ and $Q$ is represented by the pair $(z_t, s_t)$. Consequently, each transition must store this pair for the bootstrap state so that the actor target at $s_{t+2n}$ can be evaluated, together with the committed region of that next state, which becomes available only one decision step later during rollout. Algorithm~\ref{alg:training} summarizes the complete training loop described in the following two paragraphs: an asynchronous rollout process that generates transitions and an off-policy optimization process that consumes them, with both operating concurrently over a shared replay buffer.

\paragraph{The critic.}
The critic is an MLP with LayerNorm~\cite{layernorm} applied to every hidden layer. It takes $[z_t, s_t, \tilde a_{[0,n)}, a_{[n,2n)}]$ as input and predicts a scalar value. The target construction follows TD3~\cite{td3}: the target actor's action chunk is perturbed with clipped Gaussian noise, followed by a pessimistic estimate over the resulting target values. Three aspects, however, depart from the original skeleton. First, the action chunk is explicitly partitioned into the committed and execution regions, with both provided to the critic as inputs. Second, the backup follows the chunk-skipping formulation rather than a single-decision backup. Third, the fixed pair of $Q$ networks is expanded into a REDQ-style~\cite{redq} ensemble of $N$ independent critics, with the minimum taken over a randomly sampled subset of target critics rather than over the full pair.

\paragraph{The actor.}
The actor is an MLP of the same architecture. It takes $[z_t, s_t, \bar{a}_{t:t+2n}]$ as input and outputs an action chunk $a_{t:t+2n}$ of length $2n$, as illustrated in Figure~\ref{fig:architecture}. In the residual mode used in this work, the network predicts a bounded correction to the reference action, which is added to the reference chunk to produce the final action chunk. The actor loss consists of three terms, defined below for the actor output $a=\pi_\theta(z,s,\bar a)$. The first two losses follow the RLT skeleton: a $Q$-maximization term regularized by a BC anchor in the spirit of TD3+BC~\cite{td3bc}, where the anchor is defined against the reference action rather than a dataset action. Specifically, these correspond to the $Q$ term introduced in \S\ref{sec:m3} and a per-frame MSE loss against a target action $a^{\text{target}}$. The third term is specific to our framework: a smoothness regularizer that instantiates the executable set $\mathcal{E}$ introduced in \S\ref{sec:m3}. We characterize $\mathcal{E}$ by per-frame bounds on velocity, acceleration, and jerk, and relax the corresponding membership constraint into a penalty with overall weight $w_{\text{smooth}}$ and relative weight $w_k$ for each derivative order:

\begin{equation}
\mathcal L_{\text{actor}} = - Q\big(z,\ s,\ \mathrm{sg}[\tilde a_{[0,n)}],\ a_{[n,2n)}\big)
\;+\; w_{\text{bc}} \cdot \big\|a - a^{\text{target}}\big\|^2_{[0,2n)}
\;+\; w_{\text{smooth}} \sum_{k=1}^{3} w_k \cdot \big\|\Delta^k a\big\|^2_{[0,2n)},
\label{eq:actor_loss}
\end{equation}

where $\mathrm{sg}[\cdot]$ denotes the stop-gradient operator: the full action chunk is provided to $Q$, while gradients with respect to $\theta$ propagate only through the execution region.
\input{sessions/alg_training}

\paragraph{The replay buffer.}
The replay buffer is initialized by rolling out the base policy alone under the deployment loop of \S\ref{sec:m1}, with the attached network inactive; subsequent transitions collected by the asynchronous rollout are appended to the same buffer. This initialization ensures that every transition is collected under the timing assumptions of the objective, rather than from teleoperated demonstrations, and provides the critic with an initial fit to the behavior from which policy improvement begins. The reward $r$ introduced in \S\ref{sec:m2} is sparse: it is assigned only at the end of each episode, with success determined by an operator observing the rollout.

\paragraph{The update schedule.}
Updates are paced by rollout progress rather than wall-clock time. Process 2 of Algorithm~\ref{alg:training} performs $G$ critic iterations for every trajectory appended by the rollout, yielding an update-to-data ratio that keeps the update rate bounded by rollout throughput, regardless of the choice of $G$. The actor and target networks are updated less frequently: one actor update and one soft target-network update are performed every $D$ critic iterations.

%% file: sessions/fig_chunk_partitioning.tex
\begin{figure}[t]
\centering
\definecolor{astripurple}{RGB}{119, 70, 255}
\scalebox{0.88}{%
\begin{tikzpicture}[
    frame/.style={draw, rectangle, minimum height=0.5cm, minimum width=0.62cm},
    committed/.style={frame, draw=black!55, fill=white},
    execution/.style={frame, draw=astripurple!85!black, fill=astripurple!22},
    discarded/.style={frame, densely dashed, draw=gray, fill=gray!8},
    rowlab/.style={left, font=\small},
    tick/.style={font=\scriptsize},
]

\node[font=\small\bfseries, anchor=west] at (-1.6, 2.1)
    {(a) latency fluctuates: $d_k$ and $c_k$ vary from chunk to chunk};

\node[rowlab] at (-0.35, 1.25) {chunk $k\!-\!1$};
\foreach \i in {0,...,2}  { \node[execution] at (0.62*\i, 1.25) {}; }
\node[discarded] at (0.62*3, 1.25) {};
\node[font=\scriptsize, text=gray] at (0.62*4, 1.25) {$\cdots$};
\node[discarded] at (0.62*5, 1.25) {};

\node[rowlab] at (-0.35, 0) {chunk $k$};
\foreach \i in {0,...,2}  { \node[committed]     at (0.62*\i, 0) {}; }
\foreach \i in {3,...,6}  { \node[execution] at (0.62*\i, 0) {}; }
\node[discarded] at (0.62*7, 0) {};
\node[font=\scriptsize, text=gray] at (0.62*8.5, 0) {$\cdots$};
\node[discarded] at (0.62*10, 0) {};

\draw[decorate, decoration={brace, amplitude=4pt}] (-0.31, 0.32) -- (1.55, 0.32)
    node[midway, above=3pt, tick] {committed ($d_k$)};
\draw[decorate, decoration={brace, amplitude=4pt}] (1.86, 0.32) -- (4.03, 0.32)
    node[midway, above=3pt, tick] {execution ($c_k$)};
\draw[decorate, decoration={brace, amplitude=4pt}] (4.34, 0.32) -- (6.51, 0.32)
    node[midway, above=3pt, tick] {discarded};

\node[rowlab] at (-0.35, -1.45) {chunk $k\!+\!1$};
\foreach \i in {3,...,6}  { \node[committed]     at (0.62*\i, -1.45) {}; }
\draw[gray, thin] (1.55, -1.17) -- (1.55, -1.03) node[above, tick, text=black] {prediction start};
\draw[gray, thin] (4.03, -1.17) -- (4.03, -1.03) node[above, tick, text=black] {prediction end};
\foreach \i in {7,...,8} { \node[execution] at (0.62*\i, -1.45) {}; }
\node[discarded] at (0.62*9, -1.45) {};
\node[font=\scriptsize, text=gray] at (0.62*10, -1.45) {$\cdots$};

\draw[decorate, decoration={brace, amplitude=4pt, mirror}] (1.55, -1.77) -- (4.03, -1.77)
    node[midway, below=3pt, tick] {$d_{k+1}$};
\draw[decorate, decoration={brace, amplitude=4pt, mirror}] (4.03, -1.77) -- (5.27, -1.77)
    node[midway, below=3pt, tick] {$c_{k+1}$};

\node[font=\small\bfseries, anchor=west] at (-1.6, -2.75)
    {(b) the same schedule held to the budget $n$: $d_k = c_k = n$ for every $k$};

\node[rowlab] at (-0.35, -3.5) {chunk $k$};
\foreach \i in {0,...,3}  { \node[committed]     at (0.62*\i, -3.5) {}; }
\foreach \i in {4,...,7}  { \node[execution] at (0.62*\i, -3.5) {}; }
\node[discarded] at (0.62*8, -3.5) {};
\node[font=\scriptsize, text=gray] at (0.62*9, -3.5) {$\cdots$};
\node[discarded] at (0.62*10, -3.5) {};

\draw (-0.31, -3.82) -- (-0.31, -3.96) node[below, tick] {$0$};
\draw (2.17,  -3.82) -- (2.17,  -3.96) node[below, tick] {$n$};
\draw (4.65,  -3.82) -- (4.65,  -3.96) node[below, tick] {$2n$};
\draw (6.51,  -3.82) -- (6.51,  -3.96) node[below, tick] {$H$};

\node[rowlab] at (-0.35, -4.8) {chunk $k\!+\!1$};
\foreach \i in {4,...,7}  { \node[committed]     at (0.62*\i, -4.8) {}; }
\foreach \i in {8,...,11} { \node[execution] at (0.62*\i, -4.8) {}; }
\node[discarded] at (0.62*12, -4.8) {};
\node[font=\scriptsize, text=gray] at (0.62*13, -4.8) {$\cdots$};

\node[committed, minimum width=0.55cm] at (0.1, -5.7) {};
\node[right, font=\scriptsize] at (0.45, -5.7) {committed};
\node[execution, minimum width=0.55cm] at (2.5, -5.7) {};
\node[right, font=\scriptsize] at (2.85, -5.7) {execution};
\node[discarded, minimum width=0.55cm] at (4.9, -5.7) {};
\node[right, font=\scriptsize] at (5.25, -5.7) {discarded};

\end{tikzpicture}%
}
\caption{Chunk role partitioning under asynchronous execution. The committed region reuses actions already issued by the previous inference, the execution region contains the frames executed from the current chunk, and the discarded region contains frames superseded by the next inference. Chunk $k$'s committed length $d_k$ is determined by the latency of its own inference, whereas its execution length $c_k$ is determined by the latency of the subsequent inference cycle. (a) Under variable inference latency, $d_k$ and $c_k$ fluctuate independently across chunks. (b) The same schedule under a latency budget of $n$, where $d_k = c_k = n$ for every chunk, yielding a fixed execution window $[n,2n)$. Section~\ref{sec:m1} details the partition, and Section~\ref{sec:m3} defines the learning objective based on the partition.}
\label{fig:chunk_partitioning}
\end{figure}
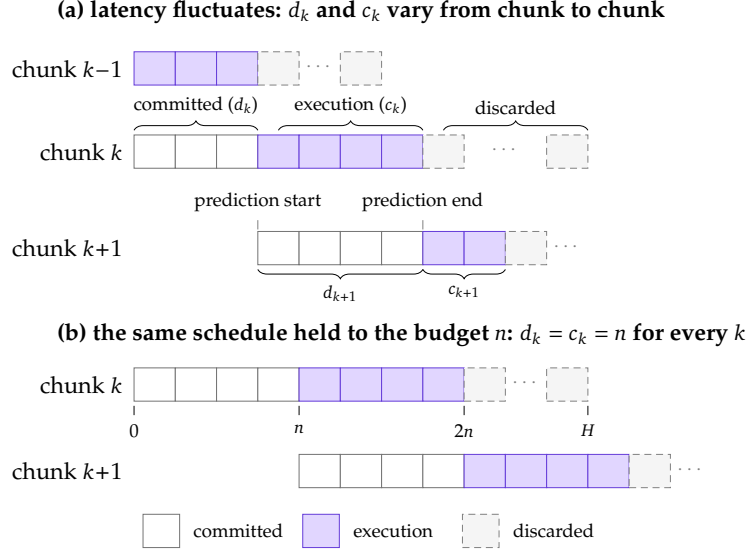

%% file: sessions/fig_architecture.tex
\begin{figure}[ht!]
\centering
\includegraphics[width=0.88\linewidth]{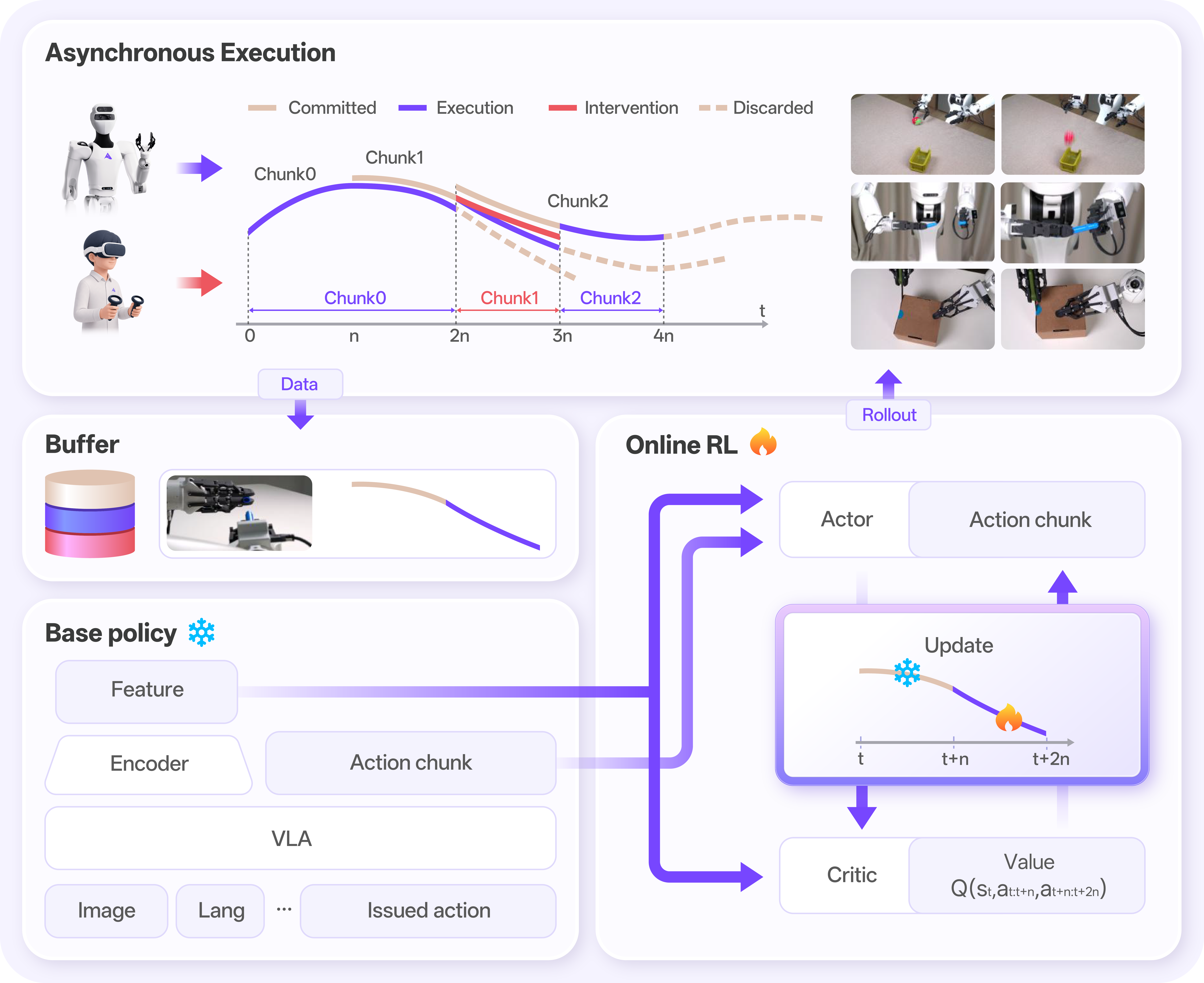}
\caption{One instantiation of our asynchronous online RL framework. The frozen base policy produces a latent representation $z_t$ and a reference action chunk $\bar{a}_{t:t+2n}$. A trainable attached network takes $[z_t, s_t, \bar{a}_{t:t+2n}]$ as input and predicts a bounded correction, which is added to the reference action to produce the final action chunk. The critic is conditioned on the corrected action chunk, while gradients are propagated to the actor only through the execution region (\S\ref{sec:m3}). Human intervention is applied to the issued action in the same raw action space, allowing intervened trajectories to be incorporated directly into training (\S\ref{sec:m4}).}
\label{fig:architecture}
\end{figure}

%% file: sessions/fig_intervention.tex
\begin{figure}[t]
\centering
\captionsetup[subfigure]{justification=centering}
\begin{subfigure}{\linewidth}
\centering
\includegraphics[width=0.7\linewidth]{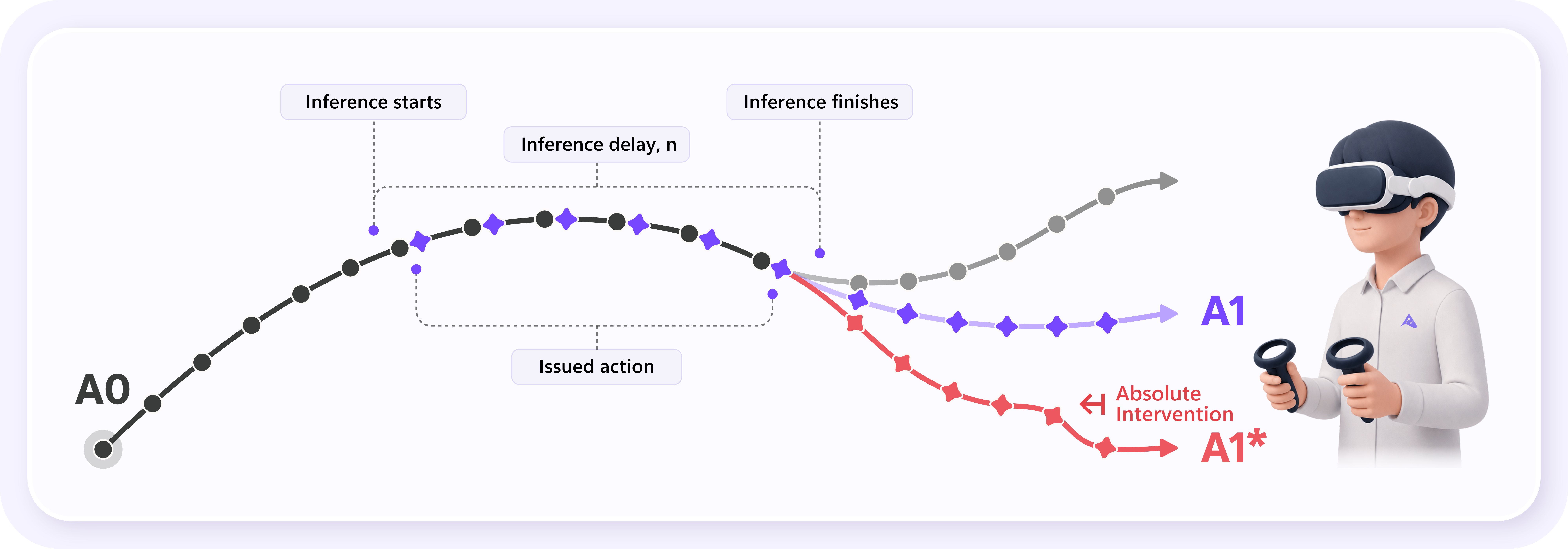}
\caption{Absolute intervention.}
\label{fig:intervention_replace}
\end{subfigure}
\\[0.8em]
\begin{subfigure}{\linewidth}
\centering
\includegraphics[width=0.7\linewidth]{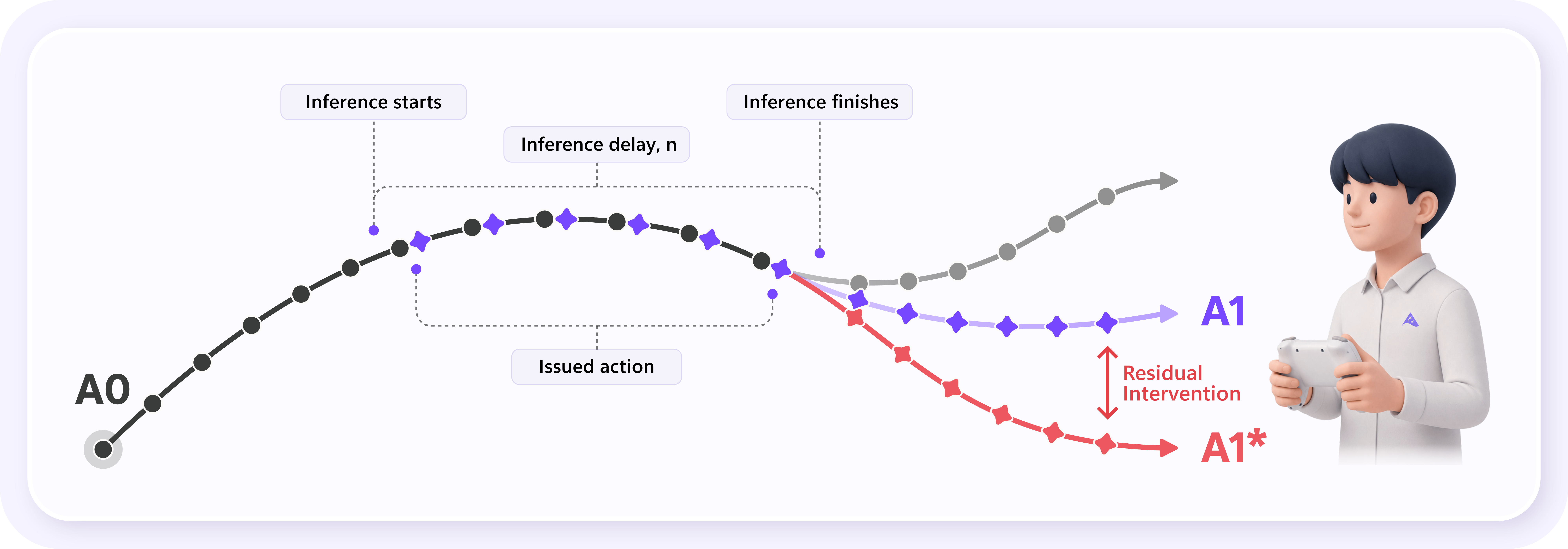}
\caption{Residual intervention.}
\label{fig:intervention_offset}
\end{subfigure}

\caption{
Two human intervention modes during policy deployment. Under \subref{fig:intervention_replace}, the teleoperated action chunk is issued directly and the actor output is discarded. Under \subref{fig:intervention_offset}, the human input is added to the actor output, allowing the operator to modify the policy-generated chunk. Both modes are treated identically downstream: the action actually issued is recorded in $\tilde a_{[0,n)}$ and used as the BC target.
}
\label{fig:intervention}
\end{figure}

%% file: sessions/alg_training.tex
\begin{algorithm}[!t]
\caption{\model training in the asynchronous inference loop}
\label{alg:training}
\begin{algorithmic}[1]
\Require frozen $\pi_{\text{base}}$ with RL-token encoder $E$, budget $n$, discount $\gamma$, soft-update rate $\tau$, step size $\eta$, update-to-data ratio $G$, actor delay $D$
\item[\textbf{Initialize:}] $\pi_\theta$ and $\{Q_{\phi_i}\}_{i=1}^{N}$
\item[] targets $\theta^- \!\gets\! \theta$ and $\phi_i^- \!\gets\! \phi_i$
\item[] replay buffer $\mathcal D$
\item[] $\tilde a_{[0,n)} \gets$ the hold action \Comment{the robot is stationary until the first chunk arrives}
\Statex \hfill\textit{Processes 1 and 2 run concurrently, sharing $\mathcal D$ and the parameters.}\hfill\
\Statex
\State \textbf{Process 1: asynchronous rollout}
\For{$t = 0, n, 2n, \dots$}
  \State observe $s_t$
  \State $\bar a_{t:t+2n} \gets \pi_{\text{base}}(s_t, \tilde a_{[0,n)})$;\ \ $z_t \gets E(\pi_{\text{base}})$
  \State $a_{t:t+2n} \gets \pi_\theta(z_t, s_t, \bar a_{t:t+2n})$
  \State $a_{[0,n)} \gets \tilde a_{[0,n)}$ \Comment{committed region over $[t, t{+}n)$}
  \vspace{4pt}
  \State $a^{\text{target}}_{[0,2n)} \gets \bar a_{[0,2n)}$
  \vspace{2pt}
  \If{teleoperation}
    \State $a_{[n,2n)} \gets a^{\text{human}}_{[n,2n)}$ \Comment{absolute or residual, \S\ref{sec:m4}}
    \State $a^{\text{target}}_{[n,2n)} \gets a_{[n,2n)}$
  \EndIf
  \State issue $a_{[n,2n)}$
  \State record a transition $\tau_t \!=\! (s_t,\ z_t,\ \tilde a_{[0,n)},\ a_{[n,2n)},\ a^{\text{target}}_{[0,2n)})$ \Comment{finished at $t{+}2n$}
  \If{$t \geq 2n$}
    \State $\tau_{t-2n} \gets \tau_{t-2n} \cup (r_{t-2n},\ s_t,\ z_t,\ \bar a_{t:t+2n},\ a_{[0,n)})$ \Comment{reward over $[t{-}2n, t)$; next state}
    \State append $\tau_{t-2n}$ to $\mathcal D$
  \EndIf
  \State $\tilde a_{[0,n)} \gets a_{[n,2n)}$ \Comment{update committed region for next step}
\EndFor
\Statex
\State \textbf{Process 2: off-policy updates}
\For{each trajectory appended by Process 1}
\For{$G$ iterations}
  \State sample a batch $B \subset \mathcal D$
  \State \textbf{for each} $\tau \!=\! (s,\ z,\ \tilde a_{[0,n)},\ a_{[n,2n)},\ a^{\text{target}},\ r,\ s',\ z',\ \bar a',\ \tilde a'_{[0,n)}) \in B$:
  \State \quad draw $\mathcal M \subset \{1, \dots, N\}$ at random
  \State \quad $a' \gets \pi_{\theta^-}(z',\ s',\ \bar a') + \epsilon$ \Comment{$\epsilon$ clipped Gaussian}
  \State \quad $y \gets r + \gamma^{2n} \min_{i \in \mathcal M} Q_{\phi_i^-}\big(z',\ s',\ \tilde a'_{[0,n)},\ a'_{[n,2n)}\big)$
  \State $\phi_i \gets \phi_i - \eta \nabla_{\phi_i} \sum_{B} \big(Q_{\phi_i}(z,\ s,\ \tilde a_{[0,n)},\ a_{[n,2n)}) - y\big)^2$ \ for all $i$
  \If{every $D$-th iteration}
    \State $\theta \gets \theta - \eta \nabla_\theta \mathcal L_{\text{actor}}$ \Comment{Eq.~\eqref{eq:actor_loss}}
    \State $\theta^- \gets \tau\theta + (1\!-\!\tau)\theta^-$
    \State $\phi_i^- \gets \tau\phi_i + (1\!-\!\tau)\phi_i^-$ \ for all $i$
  \EndIf
\EndFor
\EndFor
\end{algorithmic}
\end{algorithm}

%% file: sessions/experiments.tex
\section{Experiments}
\label{sec:exp}
This section evaluates the \model framework on real-robot manipulation tasks. We consider three tasks spanning two complementary regimes - high-speed dynamic manipulation and high-precision bimanual manipulation - to assess the framework under the competing demands of real-time responsiveness and execution reliability. We use $\pi_{0.5}$~\cite{pi05}, fine-tuned for each task, as the frozen base policy. For each task, we compare its success rate before and after online RL fine-tuning within the \model framework.

\subsection{Platform}
\label{sec:exp_platform}
All three tasks are performed on Astribot S1, a mobile bimanual robot with 25 degrees of freedom, comprising two 7-DoF arms with parallel grippers, a 4-DoF torso, a 2-DoF head, and a 3-DoF wheeled base. For the base policy, the observation consists of three camera streams (head, left wrist, and right wrist), each processed at $224\times224$ resolution, together with the robot's proprioceptive joint state. 
The base policy issues a 31-dimensional action at $30$~Hz, comprising 9-dimensional Cartesian delta poses for the left arm, right arm, and torso end-effector, along with 2 gripper commands and 2 head-joint commands. The attached actor and critic modify only the 20 arm-action dimensions, leaving the torso and head actions unchanged.
The control frequency determines the frame rate at which the latency budget $n$ in \S\ref{sec:m1} is measured. Inference is requested at $5$~Hz, so a new action chunk arrives every $200$~ms, corresponding to a latency budget of $n=6$ frames. Each chunk therefore contributes $6$ frames to its execution region while the subsequent chunk is being inferred. The base policy is trained to predict $H=32$ frames per chunk; thus, the committed and execution regions occupy $12$ frames in total, while the remaining $20$ frames form the discarded region and are superseded by the subsequent chunk before execution.

\subsection{Task Specification}
\begin{figure}[t]
\centering
\includegraphics[width=\linewidth]{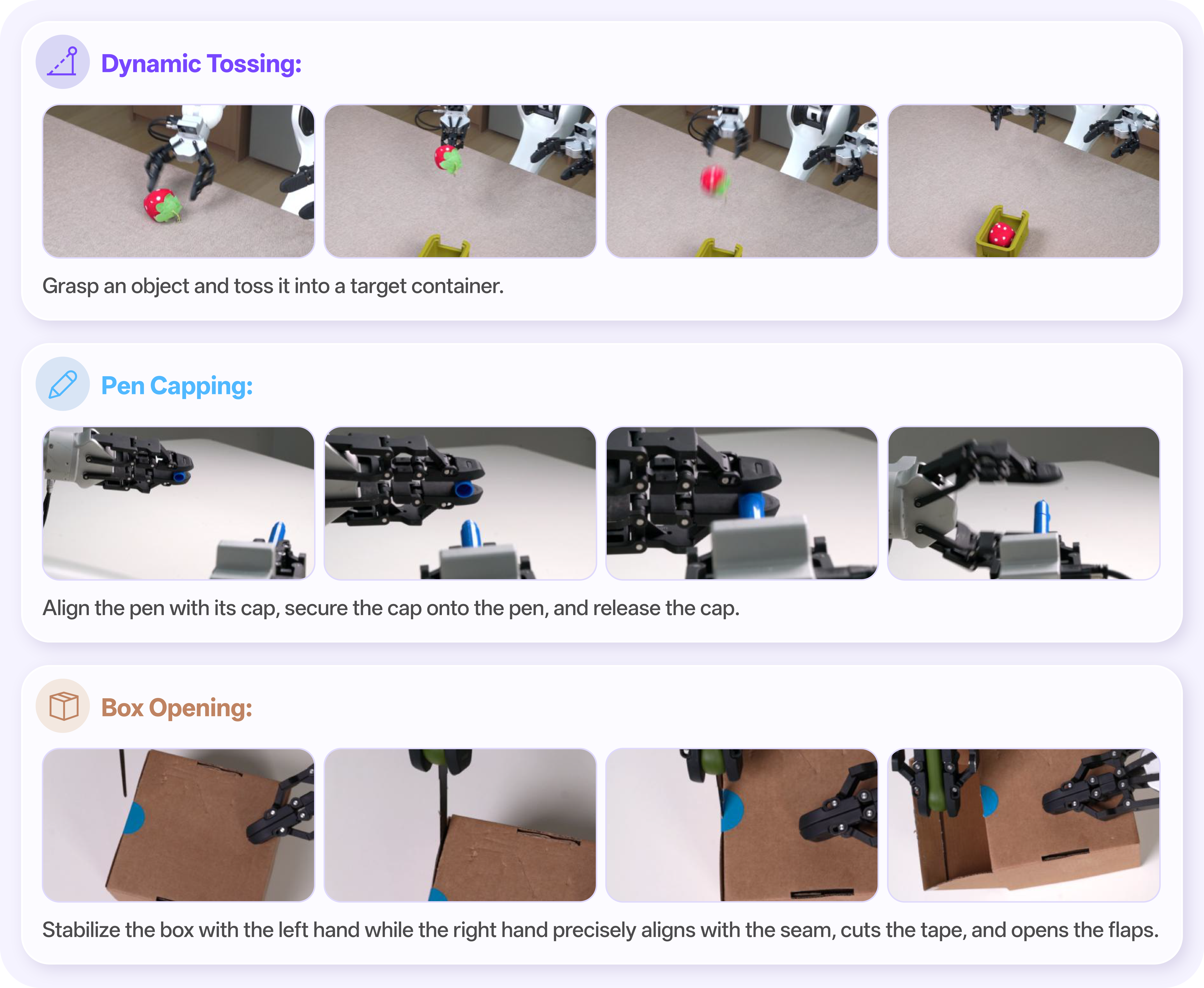}
\caption{The three real-robot tasks evaluated in this work: dynamic object tossing into a target bin, bimanual pen capping, and high-precision box opening by slitting the sealing tape and opening the flaps.}
\label{fig:tasks}
\end{figure}

\paragraph{Dynamic tossing.}
The robot is instructed to grasp a randomly placed object and toss it into a randomly placed bin. Successful tossing requires the end effector to accumulate sufficient velocity throughout the swing prior to release, rather than attaining the target velocity at a single instant. This makes the task particularly sensitive to asynchronous execution: a pause at a chunk boundary can bring the arm nearly to rest, leaving insufficient time for the remaining swing to recover the required release velocity. Consequently, successful execution requires both precise release timing and continuous motion across chunk boundaries. The bin opening is approximately $7 \times 12$~cm, while the object is approximately $6 \times 7$~cm, making the task challenging in both throwing speed and spatial accuracy.

\paragraph{Pen capping.}
The robot is instructed to align the pen with its cap, secure the cap onto the pen, and release the cap. Successful insertion requires the relative pose between the pen and cap to fall within a clearance tolerance of approximately $5$~mm. Errors beyond this tolerance cannot be reliably corrected by applying additional force, as the cap tends to deflect rather than seat properly. Since both arms are actively controlled, the insertion accuracy depends on their relative pose rather than the motion of either arm alone. This makes the task particularly challenging for the pretrained base policy: although it typically reaches the vicinity of the desired configuration, small residual pose errors can prevent successful insertion. The attached network is therefore responsible for correcting these residual errors and achieving the required insertion precision.

\paragraph{Box opening.}
The task requires the robot to stabilize a package box with its left hand while the right arm inserts a blade into the seam between the flaps and draws it along the seam to cut the tape, after which the flaps are opened. The blade is approximately $1$~mm wide, while the seam is only $2$--$3$~mm wide, requiring the insertion to locate the seam with approximately millimeter-level precision. An insertion error can fail in either direction: the blade may cut into the cardboard and damage the box, or it may miss the tape and leave the box sealed. Once the blade is inserted, the seam guides its motion, substantially reducing the positional demands during the subsequent cutting phase; the blade only needs to maintain sufficient depth to sever the tape. Thus, the primary challenge lies in precise insertion, where the pretrained policy reaches the vicinity of the desired configuration but remains insufficiently reliable.

\subsection{Training Setup}
\label{sec:exp_training}

\paragraph{Hyperparameters.}
The attached actor and critic are implemented as 3-layer MLPs with 512 hidden units per layer. We use an update-to-data ratio of $G=5$ and an actor update delay of $D=5$, such that five gradient updates are performed for each newly collected trajectory, while the actor and target networks are updated every fifth gradient step. The training batch size is set to 256. We set the correction bound to $0.05$, reflecting that the pretrained base policy is already close to successful behavior and therefore requires only small corrective adjustments. All remaining hyperparameters follow standard off-policy actor-critic practice and are kept fixed across all tasks.

\paragraph{Reward labeling and human intervention.}
The sparse reward defined in \S\ref{sec:m4} is provided by an operator through a gamepad or VR interface. An episode is terminated and assigned a binary reward only when the operator judges the task to be complete; no intermediate rewards are provided at any earlier timestep. Following each episode, the robot is returned to its initial pose and the scene is reset for the next trial. Evaluation is based solely on task success rate. During training, the operator intervenes as needed to maintain a balanced ratio of successful and failed episodes. Human-intervened actions are incorporated through the raw action space, such that they are stored and trained on in the same manner as policy-generated actions.

The two intervention modes introduced in \S\ref{sec:m4} exhibit markedly different performance on the dynamic tossing task, motivating the use of residual intervention. This difference is most pronounced for bin positions farthest from the robot, which require higher release velocities and are therefore more challenging. For these configurations, direct chunk-level teleoperation via VR achieves a success rate of approximately $30\%$, compared with approximately $80\%$ when the operator provides a residual correction to the base policy's action chunk. Direct teleoperation requires the operator to control both the direction and magnitude of the release velocity while simultaneously maintaining continuity when transitioning between human and policy control. Residual intervention instead preserves the base policy's velocity profile and requires the operator only to compensate for its deviations. This substantially reduces the control burden while better preserving motion continuity across intervention boundaries.

\subsection{Evaluation}
\label{sec:exp_eval}

\begin{figure}[t]
\centering
\includegraphics[width=\linewidth]{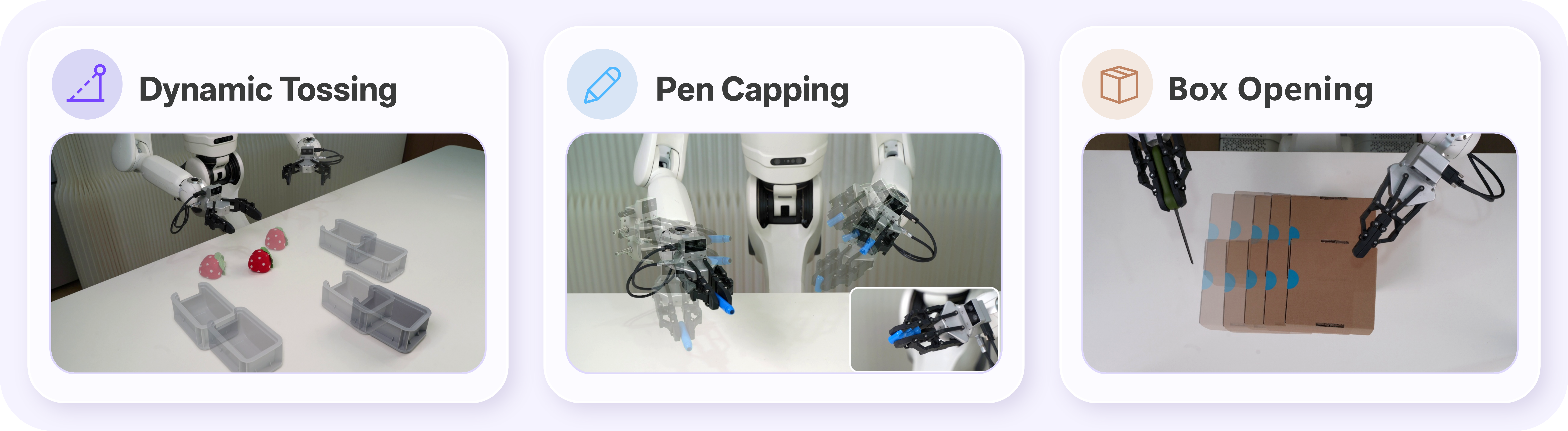}
\caption{
Initial-state configurations used for evaluation. Ghosted overlays indicate the sampled configurations. The inset illustrates the variation in the in-hand grasp pose, which is manually set during reset and is not included in the enumerated configuration combinations.
}
\label{fig:reset_randomization}
\end{figure}

\paragraph{Protocol.}
Each task is evaluated across a predefined set of initial-state configurations. Box opening uses 10 box positions, dynamic tossing uses 18 configurations comprising 3 object positions $\times$ 6 target-bin positions, and pen capping uses 12 configurations comprising 4 initial hand states $\times$ 3 pen-cap positions. Figure~\ref{fig:reset_randomization} visualizes the configurations for each task, which are identical to those used during online fine-tuning. Each configuration is evaluated once, yielding 10, 18, and 12 episodes per checkpoint for box opening, dynamic tossing, and pen capping, respectively. An episode terminates when the operator judges either that the task has been successfully completed or that it can no longer be completed. An episode is counted as successful only in the former case, and the success rate is computed as the number of successful episodes divided by the total number of evaluation episodes.

\paragraph{Comparison.}
Each task is evaluated at four points throughout the online RL process. The first is the frozen base policy $\pi_{0.5}$~\cite{pi05}, pretrained on the task and deployed with its parameters fixed, without the attached network in the control loop. This baseline represents the performance attainable through supervised pretraining alone and serves as the zero point on the horizontal axis of the learning curves shown in Figure~\ref{fig:learning_curve}. Supervised pretraining uses 1,500 teleoperated demonstrations for box opening and pen capping, and 500 for dynamic tossing. The latter requires fewer demonstrations because it is substantially more difficult to collect: across all bin positions, teleoperation succeeds on only approximately $50\%$ of attempts, as human operators face the same challenge of maintaining a continuous swing.
Online RL starts from the corresponding pretrained policy. We first collect 50 rollout episodes under the frozen base policy alone to seed the replay buffer. The attached actor and critic are then introduced into the loop, and subsequent rollout episodes are collected under the online RL procedure. The remaining three evaluation points are checkpoints from this same run, taken after 150, 200, and 250 rollout episodes. All four checkpoints are evaluated on the same hardware using the same asynchronous inference loop and latency budget. Thus, the only variable across evaluation points is the amount of online fine-tuning applied to the attached network.

\subsection{Results}

Figure~\ref{fig:learning_curve} and Table~\ref{tab:results} report the success rates of all three tasks as a function of the number of rollout episodes collected during online RL. After 250 episodes, all three tasks substantially outperform the frozen base policy: dynamic tossing improves from $39\%$ to $94\%$, pen capping from $8\%$ to $83\%$, and box opening from $30\%$ to $90\%$. The following paragraphs examine what the frozen base policy fails to capture on each task, how much of this gap is recovered through online fine-tuning, and the failure modes that remain.

\input{sessions/fig_learning_curve}

\begin{table}[t]
\centering
\begin{tabular}{ccc}
\toprule
Task & Number of Rollout Episodes & Success Rate \\
\midrule
\multirow{4}{*}{Dynamic Tossing}
  & 0 \ (base policy) & 39\% \\
  & 150 & 72\% \\
  & 200 & 83\% \\
  & 250 & 94\% \\
\midrule
\multirow{4}{*}{Pen Capping}
  & 0 \ (base policy) & 8\% \\
  & 150 & 67\% \\
  & 200 & 75\% \\
  & 250 & 83\% \\
\midrule
\multirow{4}{*}{Box Opening}
  & 0 \ (base policy) & 30\% \\
  & 150 & 20\% \\
  & 200 & 40\% \\
  & 250 & 90\% \\
\bottomrule
\end{tabular}
\caption{Success rate at each checkpoint, with the cumulative number of rollout episodes collected before evaluation. Each success rate is computed from one episode per initial-state configuration under the evaluation protocol specified in \S\ref{sec:exp_eval}.}
\label{tab:results}
\end{table}

\paragraph{Failure modes of the base policy.}
Across all three tasks, failures of the frozen base policy are systematic rather than random, primarily arising from insufficient adaptation to variations in object or target poses. In pen capping, the policy produces similar motion trajectories across different cap positions and fails to adequately adjust to the varying cap poses, resulting in only an $8\%$ success rate. In box opening, the blade exhibits a consistent leftward bias, causing it to pierce the cardboard rather than enter the tape seam. In dynamic tossing, the policy fails to appropriately modulate the release velocity with target distance: it tends to overshoot nearby bins and undershoot distant ones, while also exhibiting systematic left-right deviations for targets on opposite sides.

\paragraph{What online RL remedies.}
The attached network progressively corrects the systematic offsets of the frozen base policy, with the final checkpoint substantially improving performance across all three tasks, although the learning dynamics vary by task. Dynamic tossing and pen capping improve monotonically across checkpoints, with the largest gains occurring within the first $150$ episodes. For pen capping in particular, the success rate increases from $8\%$ to $67\%$ during this initial phase, indicating that the dominant systematic offset is largely corrected early, after which performance is primarily limited by residual variation around the corrected trajectory.
Box opening exhibits a different learning trajectory: performance initially drops to $20\%$ after $150$ episodes, falling below the frozen base policy, before recovering to $40\%$ and ultimately reaching $90\%$. This transient degradation likely reflects the exploration required for the attached network to discover effective corrections while adapting to the task.

\begin{figure}[t]
\centering
\includegraphics[width=0.8\linewidth]{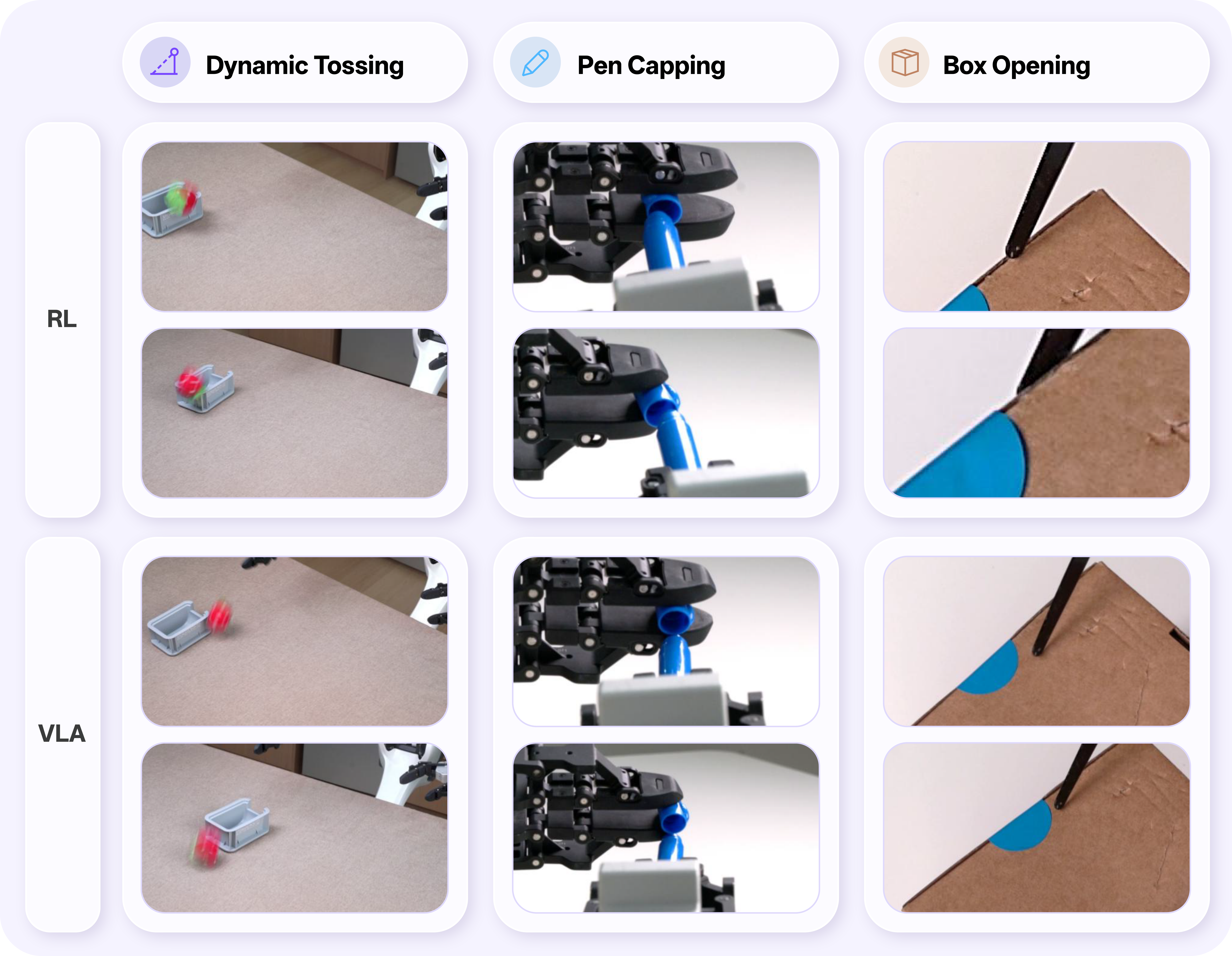}
\caption{Representative failures on dynamic tossing, pen capping, and box opening. Columns denote tasks; top and bottom rows show failures of the RL policy and frozen base policy, respectively. RL-tuned failures are near misses, while the base policy exhibits larger, systematic deviations.}
\label{fig:failures}
\end{figure}

\paragraph{Closing the last millimeter.}
Figure~\ref{fig:failures} shows representative failures on each task, at the final checkpoint (250 episodes) and under the frozen base policy for comparison. Comparing the two rows of Figure~\ref{fig:failures} shows the failures moving closer to success: the base policy misses in a consistent direction and by a visible margin, whereas what remains after RL is a small spread around the correct pose. On dynamic tossing, the object lands closer to the bin and breaks the base policy's pattern of fixed bias; on box opening, the blade's deviation from the seam is much smaller, with left-right deviations of $1$--$2$~mm around the correct position; on pen capping, the axial misalignment between cap and barrel is likewise much smaller, a small spread around the correct pose. These deviation magnitudes are precisely what makes these tasks challenging: success is not about reaching the target region but closing the last few millimeters, and an error small enough to be barely distinguishable in the image is sufficient to fail the episode.

%% file: sessions/fig_learning_curve.tex
\begin{figure}[t]
\centering
\includegraphics[width=0.72\linewidth]{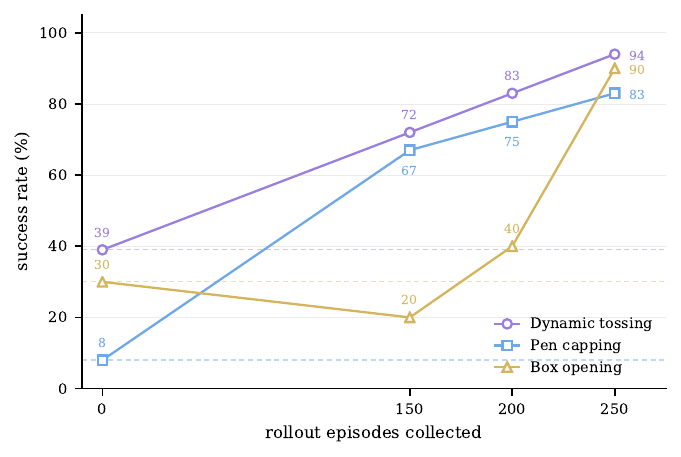}
\caption{Success rate versus online interaction episodes, with one curve per task. The zero marker denotes the frozen base policy, with its success rate extended as a dashed line for reference. Each point reports success over $N$ evaluation episodes at the corresponding checkpoint as specified in \S\ref{sec:exp}. The three nonzero checkpoints come from a single RL run per task; lines are connected for visualization only.}

\label{fig:learning_curve}
\end{figure}

%% file: sessions/conclusion.tex
\section{Conclusion, Limitations and Future Work}
\label{sec:conclusion}
This work presents \model, a framework for fine-tuning pretrained robot policies with online RL under asynchronous inference, improving success on fine-grained real-robot tasks while preserving real-time responsiveness. Asynchrony breaks the correspondence between the action optimized by the value gradient and the action actually executed by the environment. Our chunk role partitioning and gradient truncation explicitly restore this correspondence by restricting policy updates to the executed region. Operating in the raw action space further enables direct state augmentation for concurrent execution and allows human intervention to be incorporated naturally as training data.

\subsection*{Limitations}
The proposed online RL framework with asynchronous inference is general, yet the current instantiation has two known limitations:
\begin{enumerate}
  \item \textbf{Inference latency constraint.} The proposed framework operates under a fixed inference frequency, requiring each chunk-level inference to complete within a predefined latency budget. When temporal fluctuations in inference latency cause the actual computation time to exceed this budget, scheduled chunk handover cannot be guaranteed. The resultant timing misalignment destabilizes the asynchronous execution loop.
  \item \textbf{Output expressiveness bounded by the base policy.} The expressive capacity of the RL module is constrained by the base policy in two key aspects. First, the conditional features used for RL optimization are derived from the internal representations of the VLA model, so the perceptual capability of the RL branch is upper-bounded by the information richness of VLA embeddings. Second, in the current implementation, the RL module produces only bounded residual corrections upon the reference actions provided by the base policy. The correction magnitude is confined within a fixed local neighborhood, whose radius balances the effective correction range and the strength of trust-region regularization. When the base policy exhibits systematic biases on target-domain data, such local residual corrections lack sufficient expressiveness to rectify erroneous policy behaviors.
\end{enumerate}

\subsection*{Future Work}
Three broader directions emerge from this study. \textbf{First, the proposed framework may extend beyond the particular policy parameterization considered here.} Since \S\ref{sec:m3} constrains only where the value gradient is applied, the same principle could accommodate a wider class of offline-to-online learning algorithms and substantially different policy architectures. In particular, the value gradient need not be restricted to a small attached network; it could propagate through the parameters of an entire generative policy, as in QAM~\cite{qam}, whose adjoint terminal condition $\nabla_a Q$ over the generated chunk provides a natural mechanism for expressing the truncation induced by asynchronous execution. Such formulations, however, would need to treat the committed prefix explicitly: preserving previously issued actions becomes a prefix-constrained generation problem rather than an output-space overwrite. This perspective points toward a more general formulation of online policy optimization under partially committed generation.
\textbf{Second, the interaction between policy learning and action blending warrants further investigation.} While the framework rules out an external mixing operator, it does not preclude incorporating the mixture directly into the learning objective. Doing so would allow the critic to evaluate the actions that are ultimately executed, potentially providing a more faithful training signal under asynchronous deployment. This comes at the cost of a broader temporal dependency in the objective, additional continuity constraints around the blending operator, and an additional actor forward pass for each contributing chunk. More fundamentally, the blending weight typically decays toward the junction between chunks, precisely where continuity is most critical, suggesting a potentially unfavorable trade-off between execution smoothness and the strength of the learning signal. Understanding this trade-off could lead to principled formulations that jointly optimize policy improvement and temporal consistency.
\textbf{Third, scaling the evaluation to substantially broader and more diverse task distributions is essential for characterizing the fundamental expressiveness limits of the proposed adaptation mechanism.} In particular, systematically increasing the distributional shift from the pretrained base policy could reveal when local correction around the reference action ceases to be sufficient, and whether this limitation is best addressed through a more expressive adapter, a fully generative policy update, or a mechanism that can progressively relax the reference constraint. Such experiments would help distinguish limitations arising from the learning algorithm itself from those imposed by the chosen policy parameterization.